\documentclass[a4paper,fleqn]{cas-dc}

\usepackage[numbers,sort]{natbib}

\def\tsc#1{\csdef{#1}{\textsc{\lowercase{#1}}\xspace}}
\tsc{WGM}
\tsc{QE}
\tsc{EP}
\tsc{PMS}
\tsc{BEC}
\tsc{DE}
\usepackage{amsmath,amsfonts}
\usepackage{array}
\usepackage[caption=false,labelfont=sf,textfont=sf]{subfig}
\usepackage{textcomp}
\usepackage{stfloats}
\usepackage{url}
\usepackage{verbatim}
\usepackage{graphicx}
\usepackage{amsmath}
\usepackage{algpseudocode}
\usepackage{booktabs}
\usepackage{multirow}
\usepackage{amssymb}
\usepackage{algorithm}
\usepackage{cleveref}

\graphicspath{{figures/}}
\newcommand{\changes}[1]{\textcolor[rgb]{0,0,0}{#1}}

\begin{document}
\let\WriteBookmarks\relax
\def\floatpagepagefraction{1}
\def\textpagefraction{.001}
\shorttitle{Label Noise Estimation}
\shortauthors{K. Nogueira \textit{et~al.}}

\title[mode = title]{Data-Centric Benchmark for Label Noise Estimation and Ranking in Remote Sensing \changes{Binary Building} Segmentation}                      
% \tnotemark[1,2]

% \tnotetext[1]{This document is the results of the research project funded by the National Science Foundation.}
% \tnotetext[2]{The second title footnote which is a longer text matter
%    to fill through the whole text width and overflow into
%    another line in the footnotes area of the first page.}

\author[1]{Keiller Nogueira}[orcid=0000-0003-3308-6384]
\cormark[1]
\fnmark[1]
\ead{keiller.nogueira@liverpool.ac.uk}
% \ead[url]{https://www.liverpool.ac.uk/people/keiller-nogueira}
\credit{Conceptualization, Dataset, Methodology, Implementation, Experiments, Writing}
%\address[1]{, Street 129, 1043 NX Amsterdam, The Netherlands}
\affiliation[1]{organization={School of Computer Science and Informatics, University of Liverpool}, city={Liverpool}, postcode={L69 7ZX}, country={UK}}

\author[2]{Codruț-Andrei Diaconu}
\fnmark[1]
% \ead{email@}
\credit{Methodology, Implementation, Experiments, Writing, Review}
\affiliation[2]{organization={German Aerospace Center (DLR), Remote Sensing Technology Institute}, city={Wessling}, postcode={82234}, country={Germany}}

\author[3]{Dávid Kerekes}
\fnmark[1]
\credit{Methodology, Implementation, Experiments, Writing, Review}
\affiliation[3]{organization={Division of Geoinformatics, KTH Royal Institute of Technology}, city={Stockholm}, postcode={114 28}, country={Sweden}}

\author[2]{Jakob Gawlikowski}
\credit{Methodology, Implementation, Experiments, Writing, Review}
\author[2]{Cédric Léonard}
\credit{Methodology, Implementation, Experiments, Writing, Review}
\author[2]{Nassim Ait Ali Braham}
\credit{Methodology, Implementation, Experiments, Writing, Review}

\author[4]{June Moh Goo}
\credit{Dataset}
\author[4]{Zichao Zeng}
\credit{Dataset}
\affiliation[4]{organization={University College London}, city={London}, postcode={WC1E 6BT}, country={UK}}

\author[5]{Zhipeng Liu}
\credit{Dataset}
\affiliation[5]{organization={University of Exeter}, city={Exeter}, postcode={EX4 4QJ}, country={UK}}

\author[6]{Pallavi Jain}
\credit{Dataset}
\affiliation[6]{organization={Inria, Montpellier, 34000, France}, city={Paris}, postcode={75013}, country={France}}

\author[3]{Andrea Nascetti}
\credit{Writing, Review}

\author[7]{Ronny Hänsch}
\credit{Conceptualization, Writing, Review}
\affiliation[7]{organization={German Aerospace Center (DLR), Microwaves and Radar Institute}, city={Wessling}, postcode={82234}, country={Germany}}

\cortext[cor1]{Corresponding author}
% \cortext[cor2]{Equal contribution.}
\fntext[fn1]{Equal contribution.}
% \fntext[fn2]{Another author footnote, this is a very long footnote and
%   it should be a really long footnote. But this footnote is not yet
%   sufficiently long enough to make two lines of footnote text.}

% not exceed 250 words!!!!!!!!!!!!!!!!!!!!
\begin{abstract}
%Remote sensing semantic segmentation requires high-quality labels with precise spatial delineation, making it highly susceptible to annotation noise.
%This, in turn, can affect current state-of-the-art methods, severely degrading their generalization performance.
%Therefore, it is imperative to develop approaches capable of identifying such noisy labels, enabling us to deal with them.
%In this paper, we introduce three techniques for detecting noisy labels in remote sensing image segmentation datasets.
%We benchmark these approaches against traditional baselines on widely used segmentation datasets.
%% thus establishing a general and comprehensive baseline for future works.
%In each of the datasets, we demonstrate that the proposed methods outperform the baselines across various settings.
%The code is available at~\url{https://github.com/...}.
High-quality pixel-level annotations are essential for the semantic segmentation of remote sensing imagery. 
However, such labels are expensive to obtain and often affected by noise due to the labor-intensive and time-consuming nature of pixel-wise annotation, which makes it challenging for human annotators to label every pixel accurately. Annotation errors can significantly degrade the performance and robustness of modern segmentation models, motivating the need for reliable mechanisms to identify and quantify noisy training samples.
%
% This paper presents an overview of the MVEO 2024 Challenge, organized in conjunction with the BMVC 2024 workshop, which addresses the problem of ranking training samples by their label-noise level in remote sensing semantic segmentation. We describe the challenge motivation, dataset design, noise characteristics, and evaluation protocol, and provide a summary of the leading approaches submitted by participating teams. 
This paper introduces a novel data-centric benchmark, together with a new, publicly available \changes{binary} building segmentation dataset. This specific task serves as a representative and practically relevant testbed that enables controlled experimentation with different annotation perturbations. 
% This paper introduces a novel data-centric benchmark, together with a new, publicly available \changes{building segmentation} dataset\changes{, which serves as a representative and practically relevant testbed that enables controlled experimentation with different annotation perturbations. 
Furthermore, we also introduce two techniques for identifying, quantifying, and ranking training samples according to their level of label noise in remote sensing semantic segmentation.
Such proposed methods leverage complementary strategies based on model uncertainty, prediction consistency, and representation analysis, and consistently outperform established baselines across a range of experimental settings.
The outcomes of this work are publicly available at~\url{https://github.com/keillernogueira/label_noise_segmentation}.
\end{abstract}


\maketitle

\section{Introduction}

%Semantic segmentation, the task of assigning a semantic category to every pixel in an image, is crucial for remote sensing image understanding and downstream applications.
%Deep Learning, the current state of the art for this task, can achieve impressive results, especially when trained on increasingly large labeled datasets.
%However, these datasets are often created with a focus on quantity over quality.

Semantic segmentation, the task of assigning a semantic category to each pixel in an image, is a core problem in remote sensing and underpins numerous downstream applications, including land-cover mapping~\cite{fan2022land}, environmental monitoring~\cite{nogueira2024prototypical}, and urban analysis~\cite{jia2024semantic}.
Recent advances in deep learning have led to substantial performance gains, particularly when models are trained on large-scale annotated datasets~\cite{minaee2021image,zhou2024image}.
However, the effectiveness of these models is critically dependent on the quality (as well as quantity) of the training labels~\cite{roscher2024better,gong2023survey}.

Producing highly accurate pixel-level annotations is both time-consuming and expensive, and even expert-generated annotations are inherently error-prone.
As a result, many large-scale benchmark datasets are created using more scalable alternatives, such as pre-existing geographic maps, automated labeling pipelines, or crowd-sourced annotations~\cite{usmani2023towards,huang2024crowdsourcing}. 
While these approaches facilitate the creation of large and diverse datasets, they also introduce substantial levels of label noise due to different factors such as limited annotator expertise, annotation ambiguity, or systematic errors in automated pipelines~\cite{lin2022rethinking,usmani2023remote}.
Recent studies estimate that real-world datasets can exhibit noise ratios ranging from 8.0\% to 38.5\%~\cite{song2022learning}.
This issue is particularly problematic for deep learning models, which are known to memorize noisy labels, potentially leading to overfitting and degraded generalization performance~\cite{song2022learning,kimhi2024noisy}.

%Although extremely relevant, label noise detection has been more extensively studied in image classification scenarios, with relatively limited progress in image segmentation---particularly within the remote sensing domain.
%Precisely, some studies~\cite{yao2023learning,rottmann2023automated} attempt to address this issue by adapting model architectures or related components, such as regularization strategies and loss functions.
%However, these approaches are usually model-dependent and/or rely on some prior knowledge (such as access to clean validation data), therefore limiting their applicability across distinct scenarios and applications.
%Other techniques~\cite{lad2023estimating} focus on either selecting trustworthy data or correcting potentially noisy labels.

Towards tackling this, label noise has been extensively studied in image classification, where each sample is typically associated with a single label, which may be correct or incorrect~\cite{burgert2022effects}.
In contrast, semantic segmentation poses a more complex challenge: label noise is not binary but exists on a spectrum.
Within a single image, some regions of a segmentation mask may be accurate, while others are erroneous, spatially misaligned, or semantically ambiguous.
In this scenario, the structured and heterogeneous nature of noise makes its identification and quantification significantly more difficult, yet also more informative, as partially correct annotations may still contain valuable training signals.

Despite its importance, systematic approaches for identifying and quantifying label noise in semantic segmentation, particularly in remote sensing, remain underexplored.
Existing work often focuses on modifying model architectures, loss functions, or regularization strategies to improve robustness to noisy labels~\cite{xi2022nrn,zhang2023noisy,wu2023ccnr}.
While effective in specific settings, such methods are frequently model-dependent or rely on additional assumptions, such as access to clean validation data, limiting their general applicability.
% clean data or weak and dense labels
% Other approaches propose model-agnostic techniques to assign a single score to each input sample based on the pixel-wise noise level~\cite{lad2023estimating}, but they lack real-world benchmarking and applicability.
Other approaches leverage model confidence, frequently combined with specific learning strategies, such as semi- and weakly supervised learning, to down-weight or remove suspicious examples during training~\cite{liu2025cromss,otsu2025robust,xu2025mulmatch}.
While effective, these methods are typically tightly coupled to specific models or training strategies and pipelines, complicating direct comparisons across different models and datasets.

% Finally, although some of these works perform label-noise identification for image classification, to the best of our knowledge, there have been no initiatives that exploit this paradigm for remote sensing image segmentation, which is the focus of our work.

%In this paper, we introduce two approaches for detecting noisy labels in remote sensing image segmentation datasets.
%Precisely, given the images and their corresponding (potentially noisy) segmentation masks, the proposed approaches rank the examples from least to most affected by label noise.
%The main contributions of this paper are the following:
%\begin{itemize}
%    \item Introduction of three approaches for identifying noisy labels in remote sensing image datasets, thus improving the deep learning general performance; and

%    \item A full set of experiments comparing these approaches against baselines on widely used datasets, thus establishing a benchmark for future research.
%\end{itemize}

% Confident learning~\cite{northcutt2021confident} focuses on label quality by characterizing and identifying label errors in datasets, based on the principles of pruning noisy data, counting with probabilistic thresholds to estimate noise, and ranking examples to train with confidence.

To address these challenges, this paper proposes a new Data-Centric benchmark that reframes label noise estimation in semantic segmentation as a ranking problem.
% benchmark focused on ranking training samples for semantic segmentation according to their level of label noise.
Rather than making a binary decision about whether an annotation is clean or noisy, the proposed task explicitly acknowledges the continuous nature of segmentation noise by ranking images from least to most affected by pixel-wise label errors.
Such rankings, derived from model confidence and prediction-label agreement, can be used to prioritize higher-quality samples for training, data curation, or targeted manual relabeling under limited annotation budgets.

\begin{figure*}[th]
    \centering
    \includegraphics[width=\textwidth]{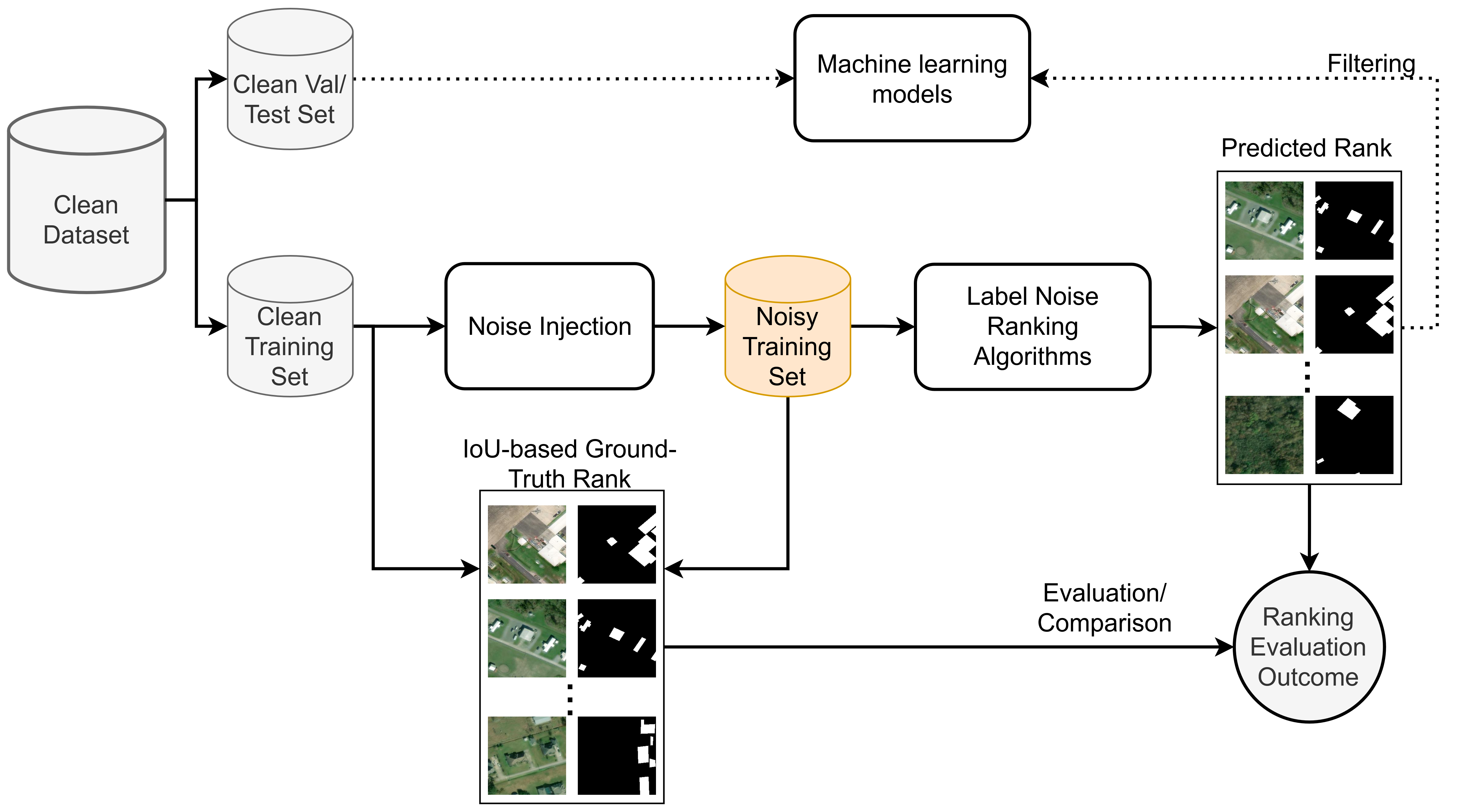}
    \caption{Overview of the proposed benchmark.
    A well-curated, clean dataset is first partitioned into training and validation/test sets. A noise injection algorithm is then applied to the training set to artificially introduce label noise. By comparing the noisy and clean annotations, a ground-truth ranking of training samples is generated based on pixel-wise Intersection-over-Union (IoU).
    The resulting noisy training set can be subsequently processed by a noise estimation method, which assigns each image a relative label-noise score, enabling a ranking from least to most affected by annotation errors.
    The predicted ranking can be evaluated against the ground-truth ranking using ranking-based metrics, such as Kendall's $\tau$~\cite{gupta2019correlation}.
    Finally, the estimated ranking can also be used to prioritize or select higher-quality samples under different data selection criteria or annotation budgets, enabling more robust model training and improved generalization, which may be assessed on the original clean validation/test sets.}
    \label{fig:overview}
\end{figure*}

This benchmark, the central outcome of the Data-Centric Land Cover Classification Challenge organized as part of the 2024 Workshop on Machine Vision for Earth Observation and Environment Monitoring (MVEO), consists of:
(i) a new, real-world, publicly available dataset comprising 5,000 training samples and 1,298 validation/test samples, and 
(ii) the two top-performing approaches from the challenge, grounded in Data-Centric and Confident Learning principles, which employ complementary strategies to estimate annotation reliability from model predictions and learned representations.
An overview of the proposed benchmark can be seen in Figure~\ref{fig:overview}.
% In introducing the proposed benchmark, we first describe the dataset and its annotation characteristics.
% We then summarize and analyze the two top-performing methods, highlighting their differing yet complementary perspectives on confidence estimation and noise ranking.
% 
Overall, by providing a unified benchmark and comparative analysis, this work aims to advance research on noise-aware learning, Confident Learning, and Data-Centric methodologies for remote sensing semantic segmentation.
% illustrates the general concept underlying the challenge: a noisy dataset is processed to estimate the relative label quality of each sample, enabling the selection of cleaner subsets for more robust model training.

% Overall, this work, an outcome of the Data-Centric Land Cover Classification Challenge of the Workshop on Machine Vision for Earth Observation and Environment Monitoring (MVEO) 2024, fills a critical gap in the literature and demonstrates the impact and importance of filtering out label noise in advancing remote sensing image segmentation.

The remainder of this paper is organized as follows: Section~\ref{sec:related_work} reviews related work, Section~\ref{sec:challenge} describes the proposed dataset, Section~\ref{sec:methods} introduces the methods, Section~\ref{sec:experimental_setup} describes the experimental setup, Section~\ref{sec:results} presents results and discussion, and Section~\ref{sec:conclusion} concludes the paper.

\section{Related work}  \label{sec:related_work}

% The impact of noisy labels, and how to deal with them, is an active field of research in Data-Centric Machine Learning~\cite{roscher2024better}.
%Most studies on noisy labels, including those in remote sensing image segmentation, focus on developing machine learning methods that are robust to the noise~\cite{zhang2020characterizing,yao2023learning,li2023semi}.
%More recently, however, label noise detection has gained increasing attention~\cite{lad2023estimating,kuan2022model,northcutt2021confident}, becoming a highly active area of research within Data-Centric Machine Learning~\cite{roscher2024better}.
%In general, existing approaches for handling noisy labels can be divided into two main categories: model recalibration and data recalibration.

Label noise has long been recognized as a major challenge in supervised learning, and a substantial body of work has focused on developing models that are robust to noisy annotations~\cite{gong2023survey,song2022learning}.
In the context of remote sensing image segmentation, most existing studies address label noise implicitly by modifying learning objectives, architectures, or training strategies to reduce sensitivity to annotation errors~\cite{kang2020robust,xi2022nrn}.
More recently, however, explicit identification and estimation of label noise has emerged as an important research direction, particularly within the paradigm of data-centric and confident learning~\cite{roscher2024better}.
Some recent works exploring this direction provide initial assessments of multiple data-centric strategies for noisy label detection, but are limited to image classification only~\cite{kroeber2026assessmentdatacentricmethodslabel}.
%focusing on image-level annotations~\cite{kroeber2026assessmentdatacentricmethodslabel}.}
In any case, rather than solely improving model robustness, these approaches aim to assess the reliability of training data itself, enabling data curation, prioritization, or targeted relabeling~\cite{lad2023estimating,northcutt2021confident}.
Overall, existing methods for handling noisy labels can be categorized into model- and data-recalibration approaches.

%Model re-calibration methods focus on developing robust machine learning approaches capable of learning effectively with the noisy labels~\cite{xi2022nrn}.
%%%%%% network modification
% Some estimate a noise matrix using specialized network architectures or loss functions.
%~\cite{zhang2023noisy} proposed a loss function to better deal with noisy labels.
%~\cite{zhang2020characterizing} combined student-teacher network with confident learning to identify and fix noisy labels.
%%%%% loss modification
% Some design loss functions that are robust to label noise.
% For example, generalized cross entropy (GCE) and symmetric cross entropy (SCE) combine both the robustness of mean absolute error and the classification strength of cross entropy loss.
%~\cite{xi2022nrn} attempted to reduce the effect of noisy labels by proposing a loss function that combines cross entropy and reverse cross-entropy (also known as symmetric cross entropy).
%%%%%%% regularization modification
% Other methods add a regularization term to prevent the network from overfitting to noisy labels

%%%%%%% 
Model re-calibration methods focus on learning robust predictors directly from noisy data by adapting loss functions, regularization schemes, or training dynamics.
Examples include noise-robust loss formulations that combine standard cross-entropy with complementary objectives to mitigate memorization of noisy labels~\cite{xi2022nrn,zhang2023noisy}, as well as approaches that introduce
explicit regularization terms to prevent the network from overfitting to noisy labels~\cite{wu2023ccnr}.
% integrate auxiliary networks or frameworks (such as the teacher–student one) to stabilize training and implicitly suppress noise effects~\cite{wang2025acoc}.
While effective under moderate noise levels, these methods often rely on strong assumptions about the noise distribution and tend to degrade when annotation noise becomes severe or highly structured~\cite{song2022learning}.

%Data re-calibration methods achieve SOTA performance by either selecting trustworthy data or correcting labels that are suspected to be noise~\cite{lad2023estimating,kuan2022model,northcutt2021confident,rottmann2023automated}.
% no remote sensing
%Some approaches filter out noisy samples or simply re-weight them to train a robust model by adding more weights to reliable samples
%~\cite{rottmann2023automated} combined connected components and network-based uncertainty~\cite{rottmann2020prediction} to identify noisy samples.
%~\cite{li2023semi} used teacher-student networks to filter out noisy samples and improve classification performance.
%~\cite{yao2023learning} used a Markov Chain to identify and iteratively fix noisy labels.
% \etal~\cite{yao2023learning} propose a novel algorithm to recover the true labels by removing the bias. The proposed algorithm requires a clean validation set, i.e., a set of well-curated annotations, to estimate and correct bias introduced by label noise.
% combine the networks with existing methods (e.g., Markov chains) to identify and remove noisy samples.
%~\cite{otsu2025robust} employed the co-training paradigm to learn multiple networks capable of identifying and filtering out noisy pixels, consequently, improving general performance.
%~\cite{liu2025cromss} used co-learning to create masks to filter out noisy pixels.

%%%%%
In contrast, data re-calibration methods explicitly target the training data by identifying, filtering, reweighting, or correcting unreliable annotations.
Several approaches estimate label confidence using model uncertainty, prediction consistency, or ensemble disagreement, and subsequently down-weight or remove suspicious samples or regions during training~\cite{liu2025cromss,lad2023estimating,aybar2023lessons,rottmann2023automated}.
% In remote sensing and dense prediction settings, uncertainty-aware pipelines have been proposed to detect noisy regions using connected components and prediction entropy~\cite{rottmann2023automated,rottmann2020prediction}.
Semi-supervised and co-training strategies have also been explored to iteratively identify trustworthy labels and suppress noisy pixels~\cite{li2023semi,otsu2025robust,xu2025mulmatch}.
Despite their effectiveness, these approaches are often tightly coupled to specific models or training pipelines, making their behavior difficult to compare across methods and datasets.

%%%%
While these approaches offer valuable insights into the behavior of noise detection methods in remote sensing settings, most of these existing methods are frequently model-dependent or rely on additional assumptions, such as access to clean validation data or weak and dense annotations, limiting their general applicability.
Moreover, they typically focus on handling noisy labels within a specific model or training pipeline, making direct comparison difficult.
Consequently, to date, there are no standardized benchmarks and evaluation protocols for assessing label-noise identification methods in remote sensing semantic segmentation, particularly from a ranking-based perspective.
%\textcolor{blue}{Another shortcoming of some prior work is its focus on image-level classification instead of dense prediction tasks.}

The present work addresses this gap by introducing a new benchmark for label noise estimation in remote sensing semantic segmentation, in which training samples are ranked according to their level of label noise.
For pixel-level label noise in semantic segmentation, annotation errors exhibit different structural properties and are typically more spatially correlated, making the problem fundamentally more complex.
By framing noise estimation as a continuous ranking task rather than a binary classification problem, this benchmark provides a unified evaluation framework for comparing diverse data-centric strategies in remote sensing semantic segmentation.

\section{Dataset} \label{sec:challenge}

In this section, we introduce:
(i) the new dataset designed for experiments on label noise in remote sensing semantic segmentation,
(ii) the noisy-label manipulations applied to generate controlled noisy training data, and
(iii) an evaluation process that enables systematic comparison of label-noise detection and ranking approaches.

% The dataset builds on the SpaceNet8 dataset~\cite{hansch2022spacenet} and provides high-resolution aerial images with building segmentation masks, containing a comprehensive and trackable amount of label noise.
% The evaluation protocol assesses how well a model ranks a given number of samples by their noise level.
% In the following, we describe the dataset, the noisy-label manipulations, and the evaluation protocol used to assess the ability of approaches to rank samples by noise level. 

%%%%%%%%%%%%%%%%%%%%%%%%%%%%%%%%%%%%%%%%%%%%%%%%%%%%%%%%%%%%%%%%%%%%
%%%%%%%%%%%%%%%%%%%%%%%%%%%%%%%%%%%%%%%%%%%%%%%%%%%%%%%%%%%%%%%%%%%%
%%%%%%%%%%%%%%%%%%%%%%%%%%%%%%%%%%%%%%%%%%%%%%%%%%%%%%%%%%%%%%%%%%%%
\subsection{Dataset} \label{sec:datasets}

%The SpaceNet8 dataset~\cite{hansch2022spacenet} consists of paired pre- and post-flooding images collected over East Louisiana and Germany.
%Each sample includes a high-resolution (0.3–0.8 m) RGB aerial image and its corresponding segmentation mask, labeled into 4 classes: flooded and non-flooded buildings and roads.
% , both downsampled to $256\times256$ pixels and a spatial resolution of 0.3-0.8m per pixel.
%For our experiments, we split the original pre-flooding images into $256\times256$ patches.
%Then, we randomly selected 5,000 samples and injected some noise into the labels (see Section~\ref{subsec:noise} for details).
%It is important to highlight that only the non-flooded building class is used in this case (i.e., the dataset is binary).
%Examples of this dataset are shown in Figure~\ref{fig:sp8_dataset}.

The proposed dataset builds on the high-resolution SpaceNet8 dataset~\cite{hansch2022spacenet}, which comprises paired pre- and post-flooding RGB images acquired by WorldView-3 over regions in East Louisiana (USA) and Germany, with spatial resolutions ranging from 0.3m to 0.8m per pixel.
Each image is accompanied by a pixel-level annotation mask containing four semantic classes: flooded buildings, non-flooded buildings, flooded roads, and non-flooded roads.
These annotations were manually annotated and subjected to rigorous quality control, and can therefore be assumed to contain minimal label noise.

In this work, only the pre-flooding images and building-related classes are considered, resulting in a binary segmentation task that distinguishes buildings from the background.
The original images are subdivided into patches of size $256 \times 256$ pixels.
From the resulting patches, 5,000 samples were randomly selected for training, while the remaining 1,298 samples were reserved for validation and testing.
Representative examples of the dataset and their corresponding clean annotations are shown in Figure~\ref{fig:sp8_dataset}.

To enable a controlled evaluation of label-noise identification and ranking, synthetic noise is injected into the segmentation masks after patch extraction.
This allows access to both clean reference labels and their noisy counterparts, which are required to quantify the accuracy of noise-level estimation.
The noise generation process is described next.
% Representative examples of the dataset, including clean and noisy annotations, are shown in Figure~\ref{fig:sp8_dataset}.

% Define image size as a new length to be able to use it in commands
\newcommand{\exFigSize}{0.127}
\newlength{\exFigDim}
\setlength{\exFigDim}{\exFigSize\textwidth}

\newcommand{\img}[1]{\includegraphics[width=\exFigDim]{#1}}
\newcommand{\rowhdr}[1]{%
  % \parbox must be positioned relative to its bottom, not its center
  \parbox[b][\exFigDim][c]{1em}{\centering\rotatebox{90}{\footnotesize#1}}%
}
\newcommand{\colhdr}[1]{%
  \parbox[c]{\exFigDim}{\centering\footnotesize #1}%
}

\begin{figure*}[!t]
    \centering
    \setlength{\tabcolsep}{3pt}
    \renewcommand{\arraystretch}{1.15}
    
    \begin{tabular}{c ccccccc}
         & \colhdr{Global\\Shrink/Expansion}
         & \colhdr{One-sided\\Shrink/Expansion}
         & \colhdr{Rotation}
         & \colhdr{Translation}
         & \colhdr{Deletion}
         & \colhdr{Vertex Addition}
         & \colhdr{False Positive\\Addition} \\
    
    \rowhdr{Image} &
    \img{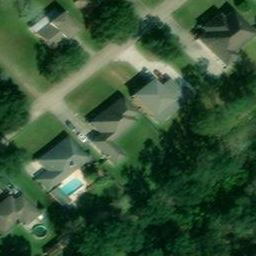} &
    \img{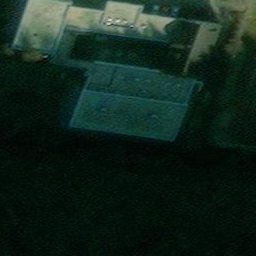} &
    \img{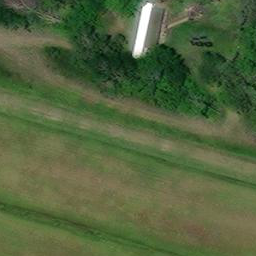} &
    \img{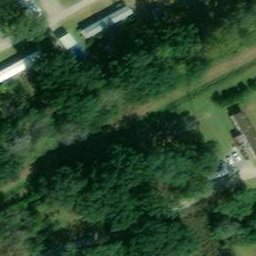} &
    \img{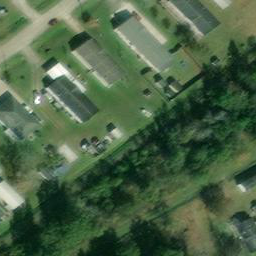} &
    \img{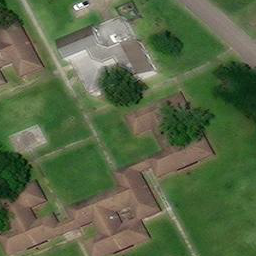} &
    \img{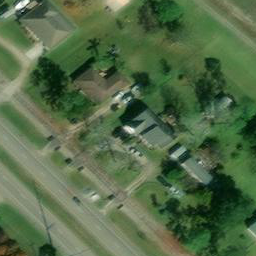} \\
    
    \rowhdr{Reference} &
    \img{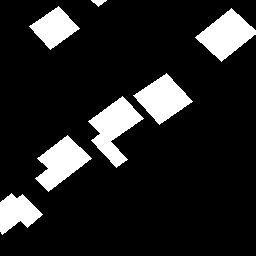} &
    \img{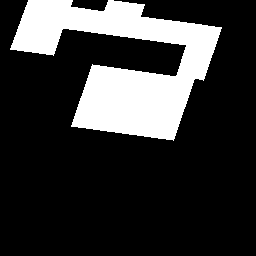} &
    \img{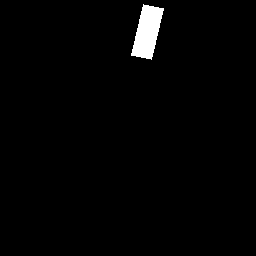} &
    \img{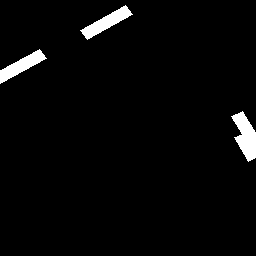} &
    \img{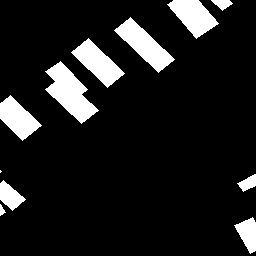} &
    \img{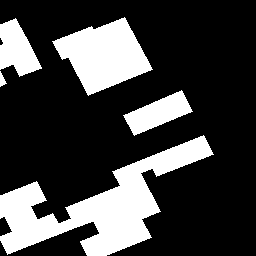} &
    \img{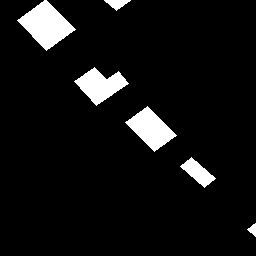} \\
    
    \rowhdr{Noisy reference} &
    \img{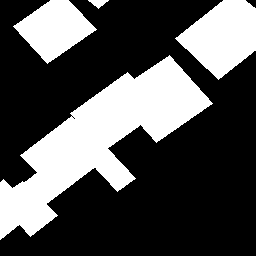} &
    \img{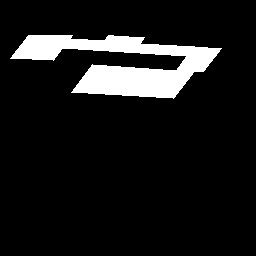} &
    \img{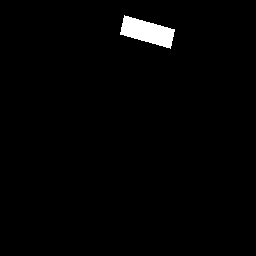} &
    \img{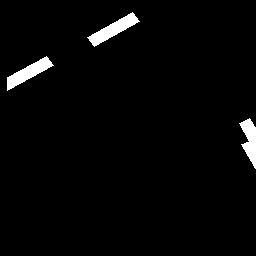} &
    \img{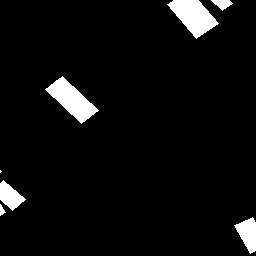} &
    \img{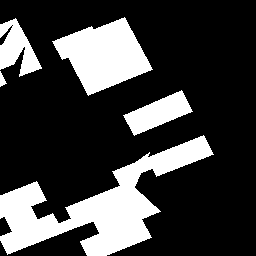} &
    \img{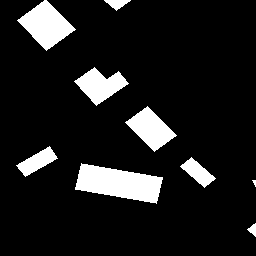} \\
    
    \end{tabular}

	\caption{Samples from the SpaceNet8 dataset~\cite{hansch2022spacenet}. The first row shows the RGB image; the second row presents the reference segmentation masks; and the third row shows the segmentation masks corrupted with synthetic noise. Each column shows a different type of noise. White pixels represent the building class whereas black pixels are the background.}
	\label{fig:sp8_dataset}
\end{figure*}

%%%%%%%%%%%%%%%%%%%%%%%%%%%%%%%%%%%%%%%%%%%%%%%%%%%%%%%%%%%%%%%%%%%%
%%%%%%%%%%%%%%%%%%%%%%%%%%%%%%%%%%%%%%%%%%%%%%%%%%%%%%%%%%%%%%%%%%%%
%%%%%%%%%%%%%%%%%%%%%%%%%%%%%%%%%%%%%%%%%%%%%%%%%%%%%%%%%%%%%%%%%%%%
\subsection{Noise Synthesis} \label{subsec:noise}

%To simulate label noise in the datasets, we synthetically corrupted the segmentation masks by applying one of seven noise types, selected at random for each sample:
%(i) global shrink/expansion – uniformly scaling all building masks by a random factor in the range $[0.3, 2.5]$,
%(ii) one-sided shrink/expansion, scales all building masks either vertically or horizontally by a random factor in the range $[0.3,  2.5]$,
%(iii) moderate rotation, rotates all building masks by a random angle in the range $[-90^\circ, 90^\circ]$,
%(vi) small translation, shifts all building masks by a random offset between $-10$ and $10$ pixels along each axis,
%(v) deletion, randomly removes a subset of building masks, with a deletion ratio between $0.3$ and $0.6$,
%(vi) vertex addition, adds between 2 and 10 random vertices to some buildings, thus altering their shapes, and
%(vii) false positive addition, inserts between 1 and 5 random building shapes (randomly cropped from other images) into the mask.
%Figure~\ref{fig:sp8_dataset} shows examples of our synthesized label noise.

To evaluate noise-level estimation, we synthetically introduced label noise into the segmentation masks of the training set.
For each sample, one of the following seven noise types was randomly applied:

\begin{enumerate}
    \item \textbf{Global shrink/expansion:} uniformly scales all building masks by a random factor in the range $[0.3, 2.5]$.
    \item \textbf{One-sided shrink/expansion:} scales all building masks along a single axis (either vertical or horizontal) by a random factor in the range $[0.3, 2.5]$.
    \item \textbf{Moderate rotation:} rotates the building masks by a randomly sampled angle in the range $[-90^\circ, 90^\circ]$.
    \item \textbf{Small translation:} shifts the building masks by a random offset of $-20$ to $20$ pixels along both the $x$ and $y$ axes.
    \item \textbf{Deletion:} randomly removes a subset of building masks, with a deletion ratio sampled from $[0.3, 0.6]$.
    \item \textbf{Vertex addition:} modifies building shapes by adding 2 to 10 vertices to the polygon contours of the buildings.
    \item \textbf{False positive addition:} inserts 1 to 5 synthetic building shapes, cropped from other images, into the mask to simulate false positives.
\end{enumerate}

These noise types were designed to reflect a range of realistic annotation errors, including geometric distortions, partial omissions, and spurious insertions.
Figure~\ref{fig:sp8_dataset} shows representative examples of the synthesized label noise.

%%%%%%%%%%%%%%%%%%%%%%%%%%%%%%%%%%%%%%%%%%%%%%%%%%%%%%%%%%%%%%%%%%%%
%%%%%%%%%%%%%%%%%%%%%%%%%%%%%%%%%%%%%%%%%%%%%%%%%%%%%%%%%%%%%%%%%%%%
%%%%%%%%%%%%%%%%%%%%%%%%%%%%%%%%%%%%%%%%%%%%%%%%%%%%%%%%%%%%%%%%%%%%
\subsection{Evaluation Process} \label{sec:ranking}

To construct the ground-truth ranking used for evaluation, we compute the pixel-wise Intersection-over-Union (IoU) between the clean segmentation masks and their noisy counterparts for each training patch.
The resulting IoU scores are then sorted to generate a reference ranking, in which samples with higher IoU (i.e., lower noise levels) are ranked above those with lower IoU.
This ranking serves as the ground truth for evaluating the accuracy of label-noise estimation and ranking methods.

After generating such ground-truth ranking, the quality of the predicted rankings can be assessed using any standard ranking-based metrics, such as Kendall's $\tau$ and Spearman's rank correlation coefficient, among others~\cite{gupta2019correlation}.

\section{Methodology} \label{sec:methods}

This section summarizes the two top-performing approaches submitted to the challenge from which the proposed benchmark emerged. Both methods address the task of estimating the relative level of label noise in remote sensing semantic segmentation data by producing a ranking of training samples according to their susceptibility to annotation errors.

Formally, given a dataset $\mathcal{D} = \{(x_i, y_i)\}_{i=1}^N$, where $x_i\in\mathbb{R}^{H\times W \times C}$ denotes a remote sensing image and $y_i\in\mathbb{R}^{H\times W}$ is the corresponding (potentially noisy) building segmentation label, 
% Each label map contains two semantic classes: \emph{background} and \emph{building}.
% Additional details regarding the dataset composition, annotation process, and noise characteristics are provided in Section~\Cref{sec:datasets}.
% Given $\mathcal{D}$, 
the objective is not to correct or filter individual pixels, but to assign each image $x_i$ a noise-related score that reflects the overall reliability of its associated label map $y_i$.
These resulting scores induce a ranking over the dataset, which can be directly compared to the ground-truth ranking and further exploited for data curation, sample selection, or annotation prioritization.
The following subsections describe the core principles and implementation details of the two approaches.

%%%%%%%%%%%%%%%%%%%%%%%%%%%%%%%%%%%%%%%%%%%%%%%%%%%%%%%%%%%%%%%%%%%%%%%%%%%%%%%%%%%%%%%%%%%%%%%%%%%%%%%%%%%%%%%%%%%%%%%%%%%%%%%%%%
\subsection{Augmented Ensemble Ranking} \label{subsec:ml4fun}

This approach is based on the RefineNet architecture~\cite{lin2017refinenet} and utilizes pretrained weights from a model trained on the INRIA building footprint dataset~\cite{maggiori2017can}, as provided by~\cite{Palnak_building-footprint-segmentation}.
Preliminary results showed that the pretrained RefinetNet model performed poorly on the given task.
Therefore, the model was fine-tuned on the provided dataset.
% To estimate the noisiness of individual labels, we measured the differences between the model’s predictions and the provided annotations.
% The differences are then sorted to create the final ranking. 

% To mitigate the effect of label noise and encourage generalization, we applied strong data augmentations \( \mathcal{A}(\cdot) \), including both geometric (e.g., rotations, scaling) and textural transformations. The optimization objective during fine-tuning was the standard pixel-wise cross-entropy loss:

% \begin{equation}
% \mathcal{L}(\theta) = -\frac{1}{N} \sum_{i=1}^{N} \sum_{c=1}^{C} \mathbf{1}_{[y_i = c]} \log p_{\theta}(y_i = c \mid \mathcal{A}(x_i)) \label{ce_loss}
% \end{equation}

% where \( C \) is the number of classes and \( p_{\theta}(y_i = c \mid \cdot) \) is the predicted probability for class \( c \) given model parameters \( \theta \). 
% The model was trained for 30 epochs. 

% \begin{figure}[h!]
% 	\centering
%     \includegraphics[width=0.5\textwidth]{figures/ml4fun/ML4FunOverview.png}
%     \caption{Illustration of the ML4FUN approach. An ensemble of models pre-trained on the Inria dataset is trained using strong data augmentations. The ensemble prediction is then used further.}
%     \label{fig:ml4fun-overview}
% \end{figure}

To mitigate the effect of label noise and encourage generalization, we applied strong data augmentations to each training pair
$(x_i, y_i)$.
Concretely, we sampled an augmentation operator $\mathcal{A}$ and formed $(\tilde{x}_i,\tilde{y}_i)=\mathcal{A}(x_i,y_i)$, where geometric transformations (horizontal flips, $90^\circ$ rotations, and small affine rotations) were applied identically to the image and the label map.
For geometric warps, images were resampled using bilinear interpolation, while label maps were resampled using nearest-neighbor interpolation to preserve discrete class IDs; pixels mapped outside the valid field of view were padded with a constant value (background, i.e., $\tilde{y}=0$).
In addition, appearance transformations (brightness/contrast, CLAHE, sharpening) were applied to the image only.

Given binary segmentation nature of the dataset, the network outputs a single-channel building probability per-pixel
$p_{\theta}(\tilde{x}_i)(u)\in(0,1)$, with pixel-location $u\in\Omega:=\{1,..,H\}\times\{1,...,W\}$, and corresponding background probability $1-p_{\theta}(\tilde{x}_i)(u)$.
We optimize the pixel-wise binary cross-entropy loss:

\begin{multline}
\mathcal{L}(\theta)
= -\frac{1}{N}\sum_{i=1}^{N}\frac{1}{|\Omega|}\sum_{u\in\Omega}
\Big[\tilde{y}_i(u)\log p_{\theta}(\tilde{x}_i)(u) \\
+\bigl(1-\tilde{y}_i(u)\bigr)\log\!\bigl(1-p_{\theta}(\tilde{x}_i)(u)\bigr)\Big],
\label{eq:bce_loss}
\end{multline}

where $\Omega$ is the set of pixel locations and $\tilde{y}_i(u)\in\{0,1\}$ denotes the augmented building mask at pixel $u$.
The model was fine-tuned for 30 epochs.

To improve model robustness, we trained an ensemble \( \mathcal{E} = \{f_{\theta^{(1)}}, f_{\theta^{(2)}}, \ldots, f_{\theta^{(K)}}\} \) with \( K = 10 \) models. Stochasticity in the training process, such as different random initializations, augmentation sequences, and data shuffling, led to diversity across ensemble members.
Final predictions \( \hat{y} \) are then obtained using majority voting at the pixel level:

\begin{equation}
\hat{y}(x) = \operatorname{mode} \left( \left\{ f_{\theta^{(k)}}(x) \right\}_{k=1}^{K} \right)
\end{equation}

%Importantly, throughout our experiments, we observed that training on the full dataset \( \mathcal{D} \) without any hold-out validation split yielded the best overall performance.
%Since we prevent overfitting on the noisy labels by using strong augmentations, the models are likely to benefit from seeing as much data as possible. This shifts the focus to producing a robust, stable disagreement signal across all samples rather than clean generalization estimates.

To estimate the noise level of individual labels, we measured the differences between the model's predictions and the provided annotations.
Concretely, for each image, we compared the predicted building mask to the given noisy building mask using IoU. We then converted this agreement into a noise score (i.e., 1 - IoU) and sorted all images from lowest to highest estimated noisiness to form the final ranking.

% This aligns with the challenge's undisclosed reference ranking, which is obtained by measuring the IoU between each sample's hidden clean label and its provided (artificially corrupted) noisy label; samples with lower clean--noisy IoU are considered more noisy and should appear later in the ranked list.

% Figure~\Cref{fig:ml4fun-overview} shows an overview of the previously described approach. Starting from a model pre-trained on the Inria dataset, an ensemble is trained on noisy labels using strong augmentations. Finally, the ensemble prediction is compared against the noisy building mask.

%%%%%%%%%%%%%%%%%%%%%%%%%%%%%%%%%%%%%%%%%%%%%%%%%%%%%%%%%%%%%%%%%%%%%%%%%%%%%%%%%%%%%%%%%%%%%%%%%%%%%%%%%%%%%%%%%%%%%%%%%%%%%%%%%%
\subsection{Regularized Variance Ranking} \label{subsec:kth}

This solution employed a pretrained ScaleMAE\cite{reed2023scalemaescaleawaremaskedautoencoder} encoder and an UperNet~\cite{xiao2018unified} decoder trained from scratch for the task.
The decoder was first trained using ADAM~\cite{kingma2017adammethodstochasticoptimization} and a cross-entropy loss on the SpaceNet2 \cite{vanetten2019spacenetremotesensingdataset} dataset with different seeds, yielding an ensemble of $K=8$ base networks.
The provided data was split into four folds for cross-validation (CV), and two networks per CV split were fine-tuned until convergence.
To reduce overfitting on potentially noisy labels, fine-tuning was continued by progressively increasing the L2 regularization coefficient of the ADAM optimizer until a peak in the F1 score on the holdout set was observed.

For each image, the IoU between the predicted labels $\hat{y}_{i,k}$ and the noisy labels $y_i$ was computed.
The best-performing prediction \(\text{IoU}_i = \text{max}_k(\text{IoU}_{i, k})\) from the ensemble was selected for each image. The images were then ranked according to their score $S_i$ calculated by the following formula:

\[S_i = \text{IoU}_i - (0.5 - \text{IoU}_i) \times \text{avg}(\text{var}_{k}(\hat{y}_{i,k}))\]

Here, the variance \(\text{var}_{k}(\hat{y}_{i,k})\) is computed pixel-wise across all eight model predictions and then averaged over all pixels in the image.
This approach penalizes images with high IoU and high variance, whereas it assigns greater weight to images with high variance and low IoU. The rationale behind this is that high-variance regions in the predictions, particularly those with low IoU, are likely to contain label noise.

% We considered the inclusion of a hyperparameter (e.g. \(\alpha\times\text{avg}(...)\)) to adjust the variance term, but the challenge's setup would have required numerous submissions to the public leaderboard for adequate tuning, making this approach impractical.

\section{Experimental Setup} \label{sec:experimental_setup}

\subsection{Baselines} \label{sec:baseline}

The introduced techniques are compared against two traditional baseline models:
(i) CleanLab~\cite{lad2023estimating}, which estimates label noise by comparing ground-truth annotations with out-of-sample model predictions using a softmax-based confidence metric (called softmin).
To obtain out-of-sample predictions (i.e., predictions from models that did not observe the corresponding samples during training), a 5-fold cross-validation protocol is applied to the noisy training set. In each iteration, a U-Net model~\cite{ronneberger2015u} is trained and used to generate predictions for one held-out subset.
(ii) Uncertainty Quantification~\cite{aybar2023lessons,rottmann2023automated}, which employs meta-classifiers to quantify label uncertainty and identify pixel-wise annotation noise within training samples.
As this approach also relies on out-of-sample predictions, we reuse the same cross-validated networks trained for CleanLab to ensure a fair comparison.

%%%%%%%%%%%%%%%%%%%%%%%%%%%%%%%%%%%%%%%%%%%%%%%%%%%%%%%%%%%%%%%%%%%%
%%%%%%%%%%%%%%%%%%%%%%%%%%%%%%%%%%%%%%%%%%%%%%%%%%%%%%%%%%%%%%%%%%%%
%%%%%%%%%%%%%%%%%%%%%%%%%%%%%%%%%%%%%%%%%%%%%%%%%%%%%%%%%%%%%%%%%%%%
\subsection{Experimental Protocol} \label{sec:protocol}

Two different experimental protocols were designed to evaluate the proposed approaches and baselines.

The first protocol corresponds to the original challenge setting and assesses the agreement between predicted and ground-truth rankings using ranking-based metrics, including Kendall's $\tau$ and Spearman's rank correlation coefficient~\cite{gupta2019correlation}.

The second protocol evaluates the practical impact of label noise on remote sensing semantic segmentation and examines how noise identification and filtering can mitigate its effects.
To this end, a U-Net~\cite{ronneberger2015u} and a SegFormer~\cite{xie2021segformer} model are trained using only the top-ranked $25\%$, $50\%$, and $75\%$ training patches selected by each method.
Validation and testing are conducted on the remaining 1,298 clean samples, as described in Section~\ref{sec:datasets}.
In this case, model performance is measured using the F1-score, a widely adopted metric for semantic segmentation~\cite{hasan2023deep}.
% and Cohen’s Kappa Index

%%%%%%%%%%%%%%%%%%%%%%%%%%%%%%%%%%%%%%%%%%%%%%%%%%%%%%%%%%%%%%%%%%%%
%%%%%%%%%%%%%%%%%%%%%%%%%%%%%%%%%%%%%%%%%%%%%%%%%%%%%%%%%%%%%%%%%%%%
%%%%%%%%%%%%%%%%%%%%%%%%%%%%%%%%%%%%%%%%%%%%%%%%%%%%%%%%%%%%%%%%%%%%
\subsection{Implementation Details} \label{sec:impl_details}

% todo: add method training settings, lr, optimizer, etc
% The Augmented Ensemble Ranking is trained for 30 epochs with a batch size of 16, optimized with Adam optimizer with default PyTorch parameters (e.g., a learning rate of 1e-4), and binary cross-entropy as the loss function.
% Ten members are finetuned based on the pretrained RefineNet, such that the stochasticity is induced by the data shuffling and the training augmentations. 

To evaluate the impact of label noise and the effectiveness of noise-aware data selection, we employ two widely used semantic segmentation architectures: U-Net~\cite{ronneberger2015u} with a ResNet-18~\cite{he2016deep} backbone, and SegFormer~\cite{xie2021segformer}.
The former is a fully convolutional model, while the latter is based on a transformer architecture.

All proposed methods are implemented using PyTorch~\cite{PyTorchNEURIPS2019_9015}.
During training, we use the following hyper-parameters: 100 training epochs, AdamW~\cite{loshchilov2017decoupled} as optimizer, learning rate of 0.001, and batch size of 128 for U-Net~\cite{ronneberger2015u} and 64 for Segformer~\cite{xie2021segformer}, reflecting GPU memory constraints.

\section{Results and Discussion} \label{sec:results}

This section presents and discusses the results obtained by the two winning solutions of the proposed challenge.
We first analyze the ranking performance using the first experimental protocol, which directly evaluates the accuracy of label-noise estimation.
We then examine the practical impact of label noise on downstream semantic segmentation performance using the second experimental protocol.
Finally, we analyze the impact of the distinct noise types in the methods.
% It is important to note that the goal of the challenge is to rank the provided images based on their level of label noise.
% Therefore, the experiments do not distinguish between training and validation data. However, we evaluate the effects of different strategies on performance under noisy and clean labels, which were disclosed only after the challenge. 

%%%%%%%%%%%%%%%%%%%%%%%%%%%%%%%%%%%%%%%%%%%%%%%%%%%%%%%%%%%%
%%%%%%%%%%%%%%%%%%%%%%%%%%%%%%%%%%%%%%%%%%%%%%%%%%%%%%%%%%%%
%%%%%%%%%%%%%%%%%%%%%%%%%%%%%%%%%%%%%%%%%%%%%%%%%%%%%%%%%%%%
\subsection{Ranking Results}

This section presents the ranking results obtained by the introduced approaches and the tested baselines.
Precisely, Table~\ref{tab:performance_metrics} summarizes the rank-based evaluation results using Kendall's $\tau$ and Spearman's correlation coefficient.
In addition to the introduced approaches and the baseline methods described previously, a \emph{Random} ranking experiment is included to provide context for the obtained results.
In this setting, samples are assigned random scores between 0 and 1.0, serving as a lower-bound reference for comparison.

% Besides the Kendal's tau ($\tau$), which represents the ranking of image-wise noise levels and was used to evaluate the challenge, we also present performance on the clean labels.
% The clean labels, i.e. the labels that are not affected by label noise, were disclosed only after the challenge for further evaluations of the approaches. To evaluate performance on clean labels, intersection over union (IoU), and pixel-wise accuracy (ACC), precision, recall, and F1 score are considered. 
% Regarding the noise ranking and the predictive performance with clean labels, one can see that the combination of pre-trained networks and fine-tuning with a high level of augmentations clearly outperforms the alternative, with $\tau=0.616$ compared to $\tau=0.512$ and IOU of 0.649 compared to 0.553.

As expected, all evaluated methods outperform the Random baseline by a substantial margin, confirming that they are able to capture meaningful information related to label noise.
More importantly, both introduced approaches significantly outperform the baseline methods, demonstrating their effectiveness in estimating and ranking the relative level of pixel-wise annotation noise in remote sensing semantic segmentation data.

Among all methods, \emph{Augmented Ensemble Ranking} achieves the strongest overall performance, attaining a Kendall's $\tau$ of 0.61 and a Spearman's rank correlation of 0.77.
This is followed by \emph{Regularized Variance Ranking}, which achieves a Kendall's $\tau$ of 0.56 and a Spearman's correlation of 0.73.
These results indicate that the proposed strategies produce outcomes that are well aligned with the ground-truth ranking, with particularly strong agreement in terms of rank correlation.

\begin{table}[tbp]
\centering
\caption{Results with respect to the produced rankings.}
\resizebox{\columnwidth}{!}{
\begin{tabular}{@{}lrr@{}}
\toprule
\multicolumn{1}{c}{\textbf{Method}} & \multicolumn{1}{c}{\textbf{Kendall's $\tau$}} & \multicolumn{1}{c}{\textbf{Spearman's}} \\ 
\midrule
Random (lower bound) & 0.0015  & 0.0021  \\
\midrule
CleanLab~\cite{lad2023estimating} & 0.1717 & 0.2383 \\
Uncertainty Quantification~\cite{aybar2023lessons,rottmann2023automated} & 0.1262 & 0.1383 \\
\midrule
Augmented Ensemble Ranking & 0.6104 & 0.7709 \\
Regularized Variance Ranking & 0.5683  & 0.7312 \\ 
\bottomrule
\end{tabular}
}
\label{tab:performance_metrics}
\end{table}

%%%%%%%%%%%%%%%%%%%%%%%%%%%%%%%%%%%%%%%%%%%%%%%%%%%%%%%%%%%%
%%%%%%%%%%%%%%%%%%%%%%%%%%%%%%%%%%%%%%%%%%%%%%%%%%%%%%%%%%%%
%%%%%%%%%%%%%%%%%%%%%%%%%%%%%%%%%%%%%%%%%%%%%%%%%%%%%%%%%%%%
\subsection{Noise Effect Results}

\begin{table}[]
\centering
\caption{Results (\% F1-score) of the U-Net model~\cite{ronneberger2015u} trained on subsets of varying sizes (25\%, 50\%, 75\%, and 100\%).}
\resizebox{\columnwidth}{!}{
\begin{tabular}{@{}lrllc@{}}
\toprule
\multicolumn{1}{c}{\textbf{Method}} & \multicolumn{1}{c}{\textbf{25\%}} & \textbf{50\%} & \textbf{75\%} & \textbf{100\%} \\
\midrule
Random (lower bound)  & 72.31 & 75.25 & 75.29  & - \\ \midrule
CleanLab~\cite{lad2023estimating} & 74.51 & 77.39         & 77.78  & -  \\
Uncertainty Quantification~\cite{aybar2023lessons,rottmann2023automated} & 75.35  & 77.30 & 77.85  & - \\ \midrule
Augmented Ensemble Ranking & 79.62 & 80.34 & 79.02  & - \\
Regularized Variance Ranking  & 78.93 & 79.56 & 79.09 & - \\ \midrule
Noisy Labels (oracle) & 80.27 & 80.98 & 80.10  & 78.26 \\
Random Clean (upper bound) & 82.08 & 83.15 & 84.53 & 84.93 \\ 
\bottomrule
\end{tabular}
}
\label{tab:unet_results}
\end{table}

\begin{table}[]
\centering
\caption{Results (\% F1-score) of the SegFormer model~\cite{xie2021segformer} trained on subsets of varying sizes (25\%, 50\%, 75\%, and 100\%).}
\resizebox{\columnwidth}{!}{
\begin{tabular}{@{}lrllc@{}}
\toprule
\multicolumn{1}{c}{\textbf{Method}} & \multicolumn{1}{c}{\textbf{25\%}} & \textbf{50\%} & \textbf{75\%} & \textbf{100\%} \\
\midrule
Random (lower bound)  & 73.49 & 74.64 & 75.90 & - \\ \midrule
CleanLab~\cite{lad2023estimating} & 74.62 & 76.35 & 78.77 & - \\
Uncertainty Quantification~\cite{aybar2023lessons,rottmann2023automated} & 74.35 & 76.20 & 78.39 & - \\ \midrule
Augmented Ensemble Ranking & 79.47 & 79.37 & 78.23 & - \\
Regularized Variance Ranking & 78.52 & 79.39 & 78.74 & - \\ \midrule
Noisy Labels (oracle) & 79.81 & 79.66 & 79.97 & 78.32 \\
Random Clean (upper bound) & 80.20 & 82.50 & 83.12 & 84.72 \\
\bottomrule
\end{tabular}
}
\label{tab:segformer_results}
\end{table}

This section evaluates the practical impact of label noise on remote sensing semantic segmentation and examines how noise identification and filtering can mitigate its effects.
Specifically, Tables~\ref{tab:unet_results} and~\ref{tab:segformer_results} present the obtained results of the U-Net~\cite{ronneberger2015u} and SegFormer models~\cite{xie2021segformer} when trained using different proportions of ranked training samples.
In addition to the proposed approaches and the baseline methods described earlier, three reference experiments are included to better contextualize the results:
(i) \emph{Random}, which serves as a lower-bound baseline, where training samples are randomly selected from the noisy training set without any noise-aware filtering,
(ii) \emph{Noisy Labels (oracle)}, where the ground-truth ranking is used to select training samples, providing an upper bound on performance achievable when the actual noise levels are known, and
(iii) \emph{Random Clean Labels}, which represents an idealized upper bound in which clean samples are randomly selected for training.

Several consistent observations can be drawn from both Tables.
First, training models using a reduced subset of less-noisy samples (e.g., the top 50\% ranked patches) consistently yields better performance than training on the full noisy dataset, a practice that is still common in the literature.
This highlights the detrimental impact of label noise on segmentation performance and underscores the importance of selective data usage.
Second, the results indicate the existence of a noise threshold beyond which incorporating additional training samples (despite increasing set size) leads to performance degradation due to the accumulation of noisy labels.
This trend is observed across both architectures, suggesting that the effect is model-agnostic and primarily driven by data quality.
Finally, across all training regimes and for both U-Net and SegFormer, the proposed methods consistently outperform the baseline approaches. This demonstrates their effectiveness in identifying and quantifying label noise, as well as their practical utility for guiding data selection strategies that lead to improved segmentation performance.

%%%%%%%%%%%%%%%%%%%%%%%%%%%%%%%%%%%%%%%%%%%%%%%%%%%%%%%%%%%%
%%%%%%%%%%%%%%%%%%%%%%%%%%%%%%%%%%%%%%%%%%%%%%%%%%%%%%%%%%%%
%%%%%%%%%%%%%%%%%%%%%%%%%%%%%%%%%%%%%%%%%%%%%%%%%%%%%%%%%%%%
\subsection{Noise Type Ablation}

In this subsection, we further analyze the influence of different noise types to better understand their effect on the proposed ranking strategies.
To this end, we compare the ground-truth ranking with: (i) a combined ranking obtained by averaging the rank positions of each sample assigned by the two proposed methods, (ii) the ranking generated by both baselines approaches.
The comparison is then analyzed separately for each noise type introduced in the dataset.

\begin{figure*}  % [!t]
	\centering
    \includegraphics[width=1.5\columnwidth]{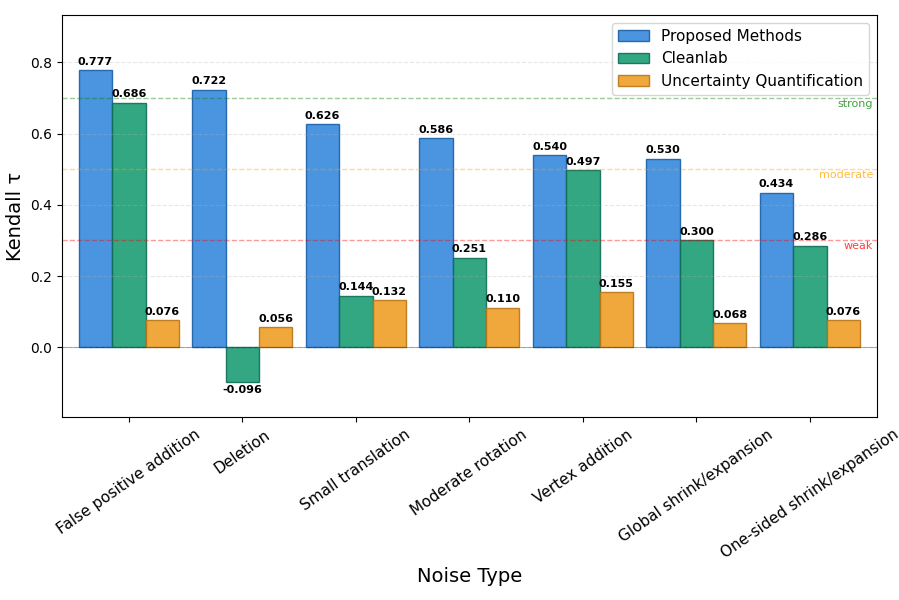}
    \caption{Comparison, in terms of Kendall's $\tau$, between the reference ranking, the average rank position assigned to each sample across the two proposed approaches, and the employed baselines}, stratified by noise type.
    \label{fig:noise_type}
\end{figure*}

Figure~\ref{fig:noise_type} reports the results in terms of Kendall's $\tau$.
The combined ranking produced by the proposed methods consistently outperformed both baseline approaches across all considered noise types.
Particularly, false-positive additions and false-negative deletions have the least impact on the combined ranking.
This indicates that both proposed methods are particularly effective at identifying and quantifying these extremely common and impactful annotation errors.
Among the baselines, CleanLab~\cite{lad2023estimating} achieved its strongest performance for false-positive additions, although it remained below the combined ranking.
In contrast, its performance deteriorated substantially for deletion perturbations, yielding the lowest correlation observed among all evaluated settings.
Similarly, the Uncertainty Quantification~\cite{aybar2023lessons,rottmann2023automated} failed to reliably identify both noise types, producing very weak correlations.
% This underscores the limitations of standard baselines when dealing with false-negative deletion noises, a gap that our approach successfully bridges.
These results indicate that the baselines struggle to reliably identify certain classes of annotation errors, whereas the proposed methods maintain robust performance across a wide range of perturbation scenarios.

Conversely, noise types involving annotation shrinking and expansion resulted in the lowest agreement between the ground-truth and the combined ranking.
A similar trend can be observed for the baseline methods, particularly for the Uncertainty Quantification approach~\cite{aybar2023lessons,rottmann2023automated}, which achieved very weak correlations under these settings.
%Under these perturbation settings, the Uncertainty Quantification method~\cite{aybar2023lessons,rottmann2023automated} also struggled, presenting very low $\tau$ scores, similar to the false-positive additions and false-negative deletions.
We attribute this behavior to the ambiguity introduced at object boundaries, where small geometric distortions can lead to large pixel-wise discrepancies while largely preserving the semantic identity of the object, making such noise patterns more challenging to detect and rank accurately.

% Moreover, global and one-sided shrink–expansion noise patterns may be inherently more difficult to detect because their visual effect can resemble perspective-induced vertical building displacements, which are frequently observed in high-resolution remote sensing imagery.

% The same can be seen in \Cref{fig:ml4fun-noisetypes_cross_eval}, where rank evaluation is performed only on a subset of label noise types: rows and columns with false positives or false negatives achieve clearly higher scores. The lowest performance, by contrast, is due to the noise types "One-sided expand" and "One-sided shrink". 

% Considering the different types of label noise, false positives (locations mislabeled as buildings) and false negatives (missing labels for actual buildings) are clearly the most problematic for ranking.
% This is most likely because the underlying segmentation model does not perform perfect segmentation, and thus other types of label noise are easily hidden in the overall noise in the prediction.
% From a practical standpoint, this may not be particularly relevant for two reasons: in tasks such as building classification, false positives and false negatives are typically the most common and of greatest practical interest.
% On the other hand, for tasks where pixel-wise exact prediction is highly relevant, the proposed approach would perform significantly better at ranking noise, as its performance directly depends on the model's performance.

\section{Conclusions} \label{sec:conclusion}

In this paper, we introduce a novel Data-Centric benchmark that reframes label noise estimation in semantic segmentation as a ranking problem.
Rather than making a binary decision about whether an annotation is clean or noisy, the proposed task explicitly acknowledges the continuous nature of segmentation noise by ranking images from least to most affected by pixel-wise label errors.
This formulation enables a more nuanced and practically relevant assessment of annotation reliability in large-scale remote sensing datasets.

As part of this benchmark, we introduce a new set of noisy building segmentation labels, derived from the SpaceNet8~\cite{hansch2022spacenet} dataset. 
The resulting dataset incorporates multiple realistic noise types and is made publicly available together with clean, noise-free ground-truth label masks, enabling controlled and reproducible evaluation of label-noise estimation methods.
In addition, we present and analyze the two winning approaches submitted to the Data-Centric Land Cover Classification Challenge from which the proposed benchmark emerged.
Both methods rely on ensembles of neural networks and exploit discrepancies between predicted and ground-truth segmentation masks as proxies for estimating annotation noise, albeit through complementary strategies.
% In particular, the combination of strong correlations, which prevented the models from overfitting to label noise, led to convincing results.

Experimental results demonstrate the effectiveness of the proposed techniques, which consistently outperform traditional baselines across multiple evaluation protocols.
A key empirical finding is that training models on a reduced subset of less-noisy samples (such as the top 50\% ranked patches) often yields superior performance compared to training on the full noisy dataset, a practice that remains common in the literature.
% This highlights the detrimental impact of label noise on segmentation performance and underscores the importance of selective data usage.
% These findings highlight that carefully select higher quality samples can not only improve model performance but also substantially reduce training time and computational costs.
This observation highlights the detrimental impact of label noise on segmentation performance and underscores the importance of selective data usage.
Moreover, it shows that prioritizing higher-quality samples can not only improve model accuracy but also substantially reduce training time and computational costs.

Overall, this work and its findings open several promising research directions, including:
(i) more effective prioritization of high-quality samples for training, data curation, or targeted manual relabeling under limited annotation budgets, and
(ii) the advancement of noise-aware learning, Confident Learning, and Data-Centric methodologies for remote sensing semantic segmentation through the provision of a unified benchmark and comparative evaluation framework.

For future work, several directions can be explored. While the synthetic perturbations considered in this work enable controlled and reproducible experiments, they do not fully capture the complexity of annotation errors encountered in real-world remote sensing datasets. In practice, label noise may exhibit systematic biases, spatial dependencies, annotator-specific patterns, or domain-dependent artifacts that are difficult to model explicitly. Nevertheless, the use of synthetic noise provides an important advantage: the exact type and location of annotation errors are known, allowing a rigorous and quantitative evaluation of label-noise detection performance. In contrast, evaluations on genuinely noisy datasets are challenging because reliable ground-truth information about annotation errors is typically unavailable, making an objective assessment of detection accuracy difficult. Therefore, the presented benchmark should primarily be interpreted as a controlled comparative framework for analyzing the behavior of different methods under well-defined perturbation scenarios. Investigating the transferability of these findings to real-world noisy datasets remains an important direction for future work.
Other directions of future work include to extend the benchmark to additional semantic classes and remote sensing modalities and further investigate the relationship between noise characteristics and model uncertainty in semantic segmentation tasks.

\printcredits

\section*{Aknowledgements}

The work of CAD, JG, CL, and NB was supported by the Helmholtz Association's Initiative and Networking Fund on the HAICORE (at) FZJ partition.

% \bibliographystyle{cas-model2-names}
\bibliographystyle{elsarticle-num}
\bibliography{references}

\vskip3pt

%%%%%%%%%%%%%%%%%%%%%%%%%%%%%%% BIOGRAPHIES
\bio{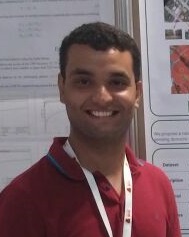}
Keiller Nogueira received the BSc degree in Computer Science from Universidade Federal de Vi\c{c}osa (Brazil) in 2012. He has received the MSc and PhD degrees in Computer Science from the Universidade Federal de Minas Gerais, Brazil, in 2015 and 2019, respectively.
Since 2024, he has been a lecturer in the School of Computer Science and Informatics at the University of Liverpool, UK. Before that, from 2019 to 2024, he was a lecturer in Data Science at the University of Stirling, UK.
He has published several high-quality articles in leading journals and conferences.
His research interests include Deep and Machine Learning, Pattern Recognition, Image Processing, Computer Vision, and Remote Sensing.
\endbio

\bio{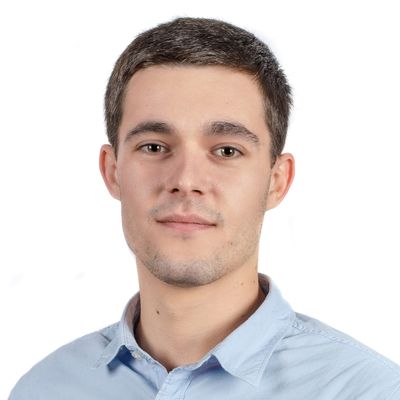}
Codruț-Andrei Diaconu received the B.Sc. degree in computer engineering from the Gheorghe Asachi Technical University of Iași, Romania, and the M.Sc. degree in computing science (Data Science and Systems Complexity) from the University of Groningen, The Netherlands. He is currently pursuing a Ph.D. degree at the German Aerospace Center (DLR) in collaboration with the Technical University of Munich (TUM), Germany. His research focuses on deep learning for Earth observation and cryosphere remote sensing, with particular interest in multi-modal glacier monitoring.
\endbio

\bio{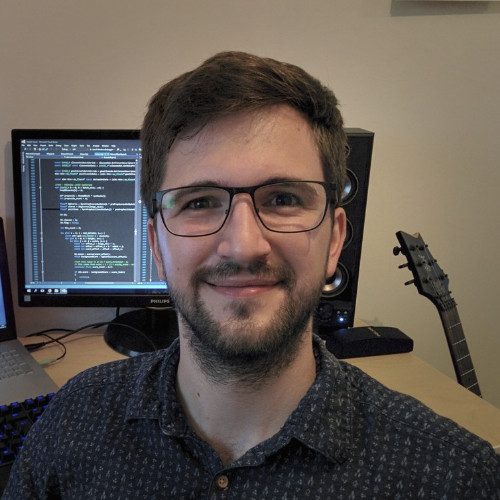}
Dávid Kerekes is a PhD student in Geoinformatics at KTH Royal Institute of Technology, focusing on the adoption of deep learning models for Earth Observation data analysis. He holds a Master’s degree in Artifical Intelligence from KU Leuven and has actively participated in several international geospatial AI challenges, including SpaceNet7, xView3, and Embed2Scale. Through these experiences, he has developed expertise in best practices for applying deep learning to large-scale EO datasets.
\endbio

\bio{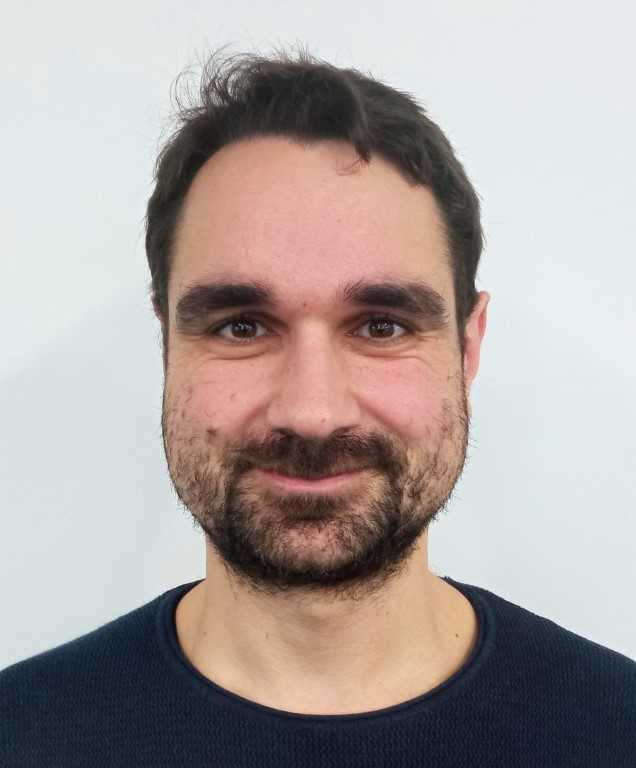}
Jakob Gawlikowski received a B.Sc. and an M.Sc. degree in Mathematics from the Technical University of Munich (TUM), Germany in 2016 and 2019, respectively. He is currently pursuing a PhD at the German Aerospace Center (DLR) in collaboration with TUM. Since 2025, he is with the DLR Remote Sensing Technology Institute. His research focuses on the trustworthiness of machine learning approaches, with particular emphasis on uncertainty quantification (UQ), explainable AI (xAI), and multi-modal learning setups.
\endbio

\bio{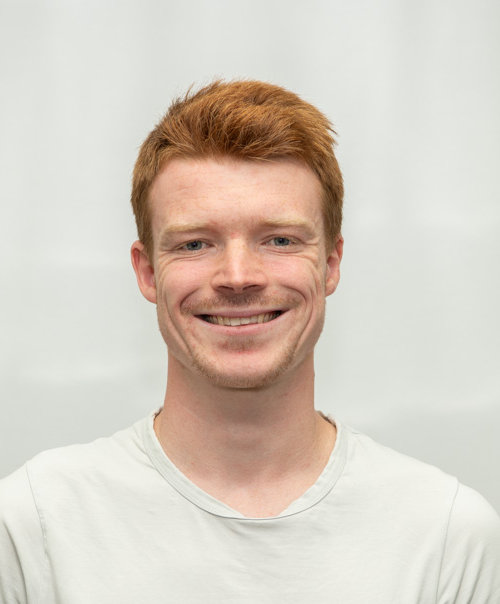}
Cédric Léonard received an engineering degree in electronics from the National Institute for Applied Sciences (INSA) Rennes, France, and an M.Sc. degree in computer science (information technology) from Åbo Akademi University, Finland. He is currently pursuing a PhD at the German Aerospace Center (DLR) in collaboration with the Technical University of Munich (TUM), Germany. His research activities focus on Deep Learning for Earth Observation and Synthetic Aperture Radar (SAR), with particular interest in efficient AI approaches for onboard and edge computing, particularly FPGA-based implementations.
\endbio

\bio{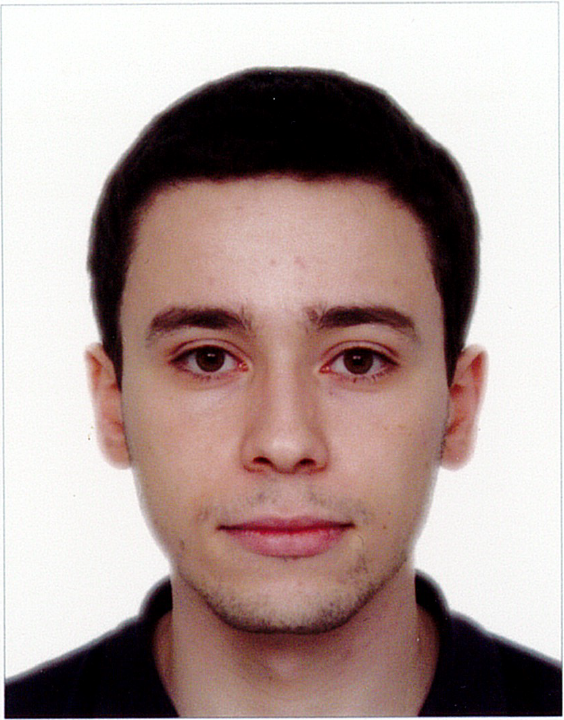}
Nassim Ait Ali Braham received his M.Sc. degree in computer science from Ecole nationale Supérieure d'Informatique (ESI), Algiers, Algeria, in 2019 and a M.Sc. degree in Artificial Intelligence and Data Science from Université Paris Dauphine-PSL, Paris, France, in 2020. He is pursuing a Ph.D. degree at the German Aerospace Center, Wessling, Germany, and at the Technical University of Munich, Munich, Germany. In 2019, he spent six months as a research intern at the LIRIS-CNRS laboratory, Lyon, France. In 2020, he spent six months at the LAMSADE-CNRS laboratory, PSL Research University, Paris, France. In 2023, he spent six months as a visiting researcher at INRIA THOTH, Grenoble, France. His research interests include deep learning, computer vision, self-supervised learning, and remote sensing.
\endbio

\bio{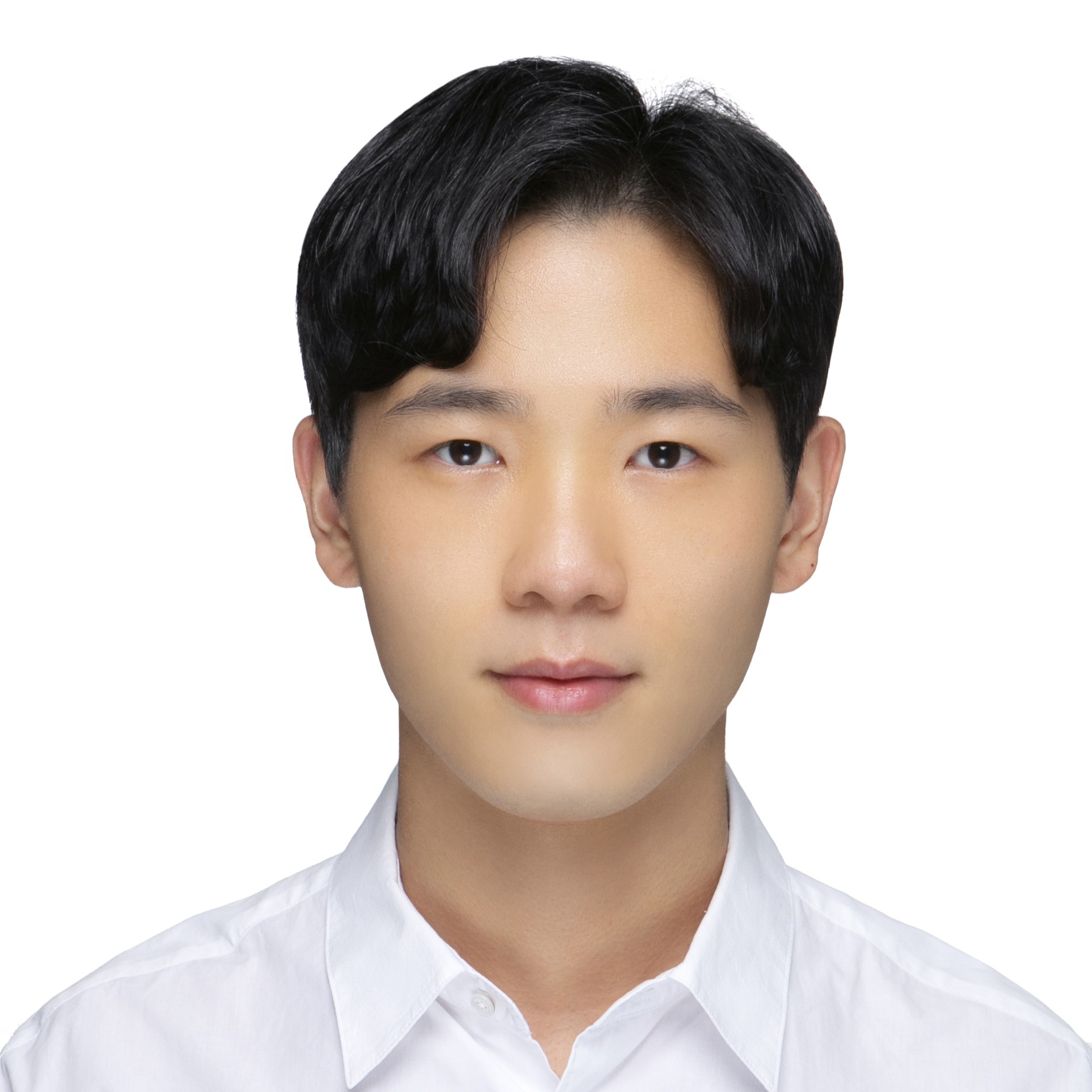}
June Moh Goo received the BEng degree in Engineering Mathematics from the University of Bristol, United Kingdom, in 2019, and the M.Sc. degree in Data Science from the University of Bath, United Kingdom, in 2020. He also undertook postgraduate studies in Machine Learning at University College London (UCL), UK. He is currently working toward the Ph.D. degree in Civil, Environmental, and Geomatic Engineering at University College London, UK. He is an active member of the Photogrammetry and 3D Imaging Group and is supported by the Engineering and Physical Sciences Research Council through an industrial CASE studentship with Ordnance Survey. His research interests include LiDAR super-resolution, domain adaptation for 3D point clouds, and label-efficient learning for 3D data.\\
\endbio

\bio{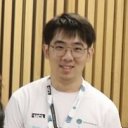}
Zichao Zeng is currently pursuing the PhD degree at University College London (UCL), UK. His research focuses on improving interactive pedestrian navigation using Computer Vision techniques, with particular emphasis on vision-based geolocalization. His doctoral research is funded by UK Research and Innovation (UKRI) through the Engineering and Physical Sciences Research Council (EPSRC), in collaboration with Ordnance Survey (OS), and supported by the UCL International Scholar Doctoral Training Award (ISAD). His research interests include Computer Vision, Visual Localization, Multimodal Deep Learning, and Robot Vision.
\endbio

\bio{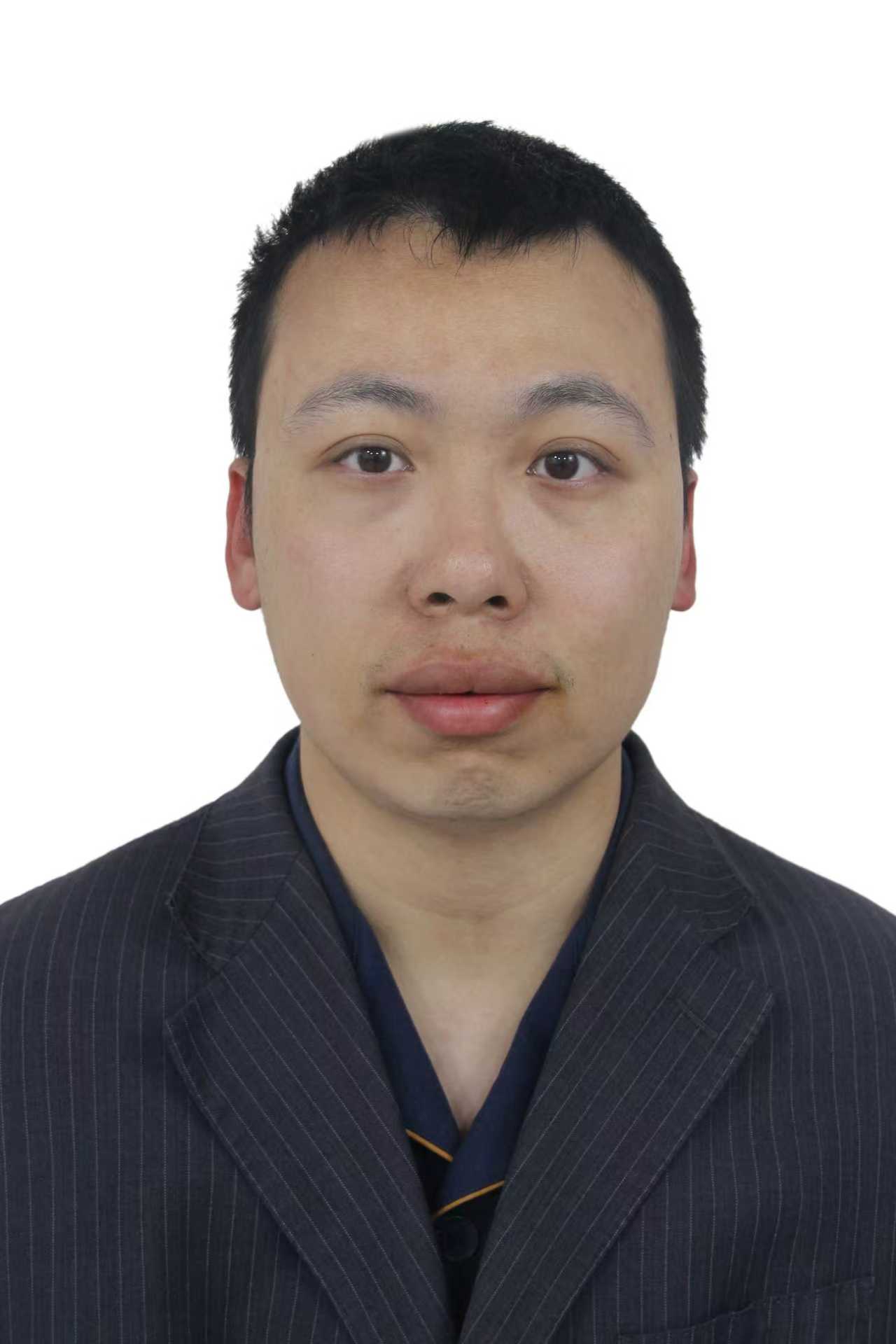}
Zhipeng Liu received the B.Eng. degree in software engineering from Harbin Institute of Technology, Weihai, China, in 2020, and the M.Sc. degree in advanced computer science from the University of Exeter, Exeter, U.K., in 2022, where he is currently pursuing the Ph.D. degree. His research interests include computer vision, remote sensing, and disaster monitoring.\\

\endbio

\bio{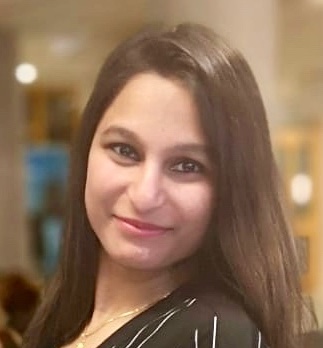}
Pallavi Jain received the M.Sc. and Ph.D. degrees in Computer Science from Technological University Dublin, Ireland, in 2018 and 2023, respectively. Since 2023, she has been a postdoctoral researcher at UMR TETIS, in collaboration with INRIA. Her research interests include computer vision, self-supervised learning, cross-modal learning, and interpretable machine learning, with particular emphasis on remote sensing applications.
\endbio

\bio{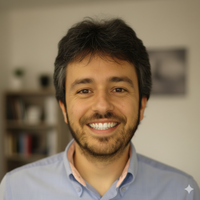}
Andrea Nascetti is Associate Professor at KTH Royal Institute of Technology, working at the intersection of computer vision and remote sensing developing ML and DL methods using multi-modal satellite time series for biomass estimation, 3D mapping, and wildfire monitoring. He received his M.Sc. and Ph.D. degrees in Geodesy and Geomatics from Sapienza University of Rome. His recent research investigates geospatial foundation model capabilities through the development of large-scale benchmarks such as PANGAEA-Bench. He also led the international data challenge (i.e. BioMassters), establishing a widely used benchmark for forest biomass estimation. 
\endbio

\bio{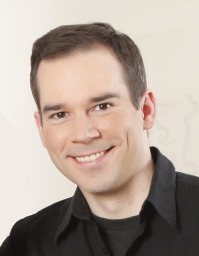}
Ronny Hänsch received his Diploma and Ph.D. in computer science from TU Berlin in 2007 and 2014. He is a senior scientist at the DLR Microwave and Radar Institute, leading the Machine Learning Team in the SAR Technology Department. He has served as chair of the GRSS IADF technical committee, editor of the GRSS eNewsletter, and co-chair of the ISPRS working group on Image Orientation and Sensor Fusion. He is editor-in-chief of IEEE GRSL and associate editor of the ISPRS Journal. He has led multiple remote sensing competitions, including IEEE Data Fusion Contests and SpaceNet 8 and 9 Challenges. 
\endbio

\end{document}